\documentclass{easychair}

\usepackage{multirow}
\usepackage{url}
\usepackage{hyperref}
\usepackage{graphicx}
\usepackage{algorithm}
\usepackage{algpseudocode}
\usepackage{booktabs}
\usepackage{wrapfig}
\usepackage[font=small,labelfont=bf]{caption}

\title{Thinking Tokens for Language Modeling}
\author{David Herel, Tomas Mikolov}
\institute{
   Faculty of Electrical Engineering, Czech Technical University in Prague \\
  Czech Institute of Informatics, Robotics and Cybernetics, Czech Technical University in Prague \\
  \email{hereldav@fel.cvut.cz}
 }
\authorrunning{David Herel, Tomas Mikolov}
\titlerunning{Thinking Tokens for Language Modeling}
\begin{document}

\maketitle
\begin{abstract}
How much is 56 times 37? Language models often make mistakes in these types of difficult calculations. This is usually explained by their inability to perform complex reasoning. Since language models rely on large training sets and great memorization capability, naturally they are not equipped to run complex calculations. However, one can argue that humans also cannot perform this calculation immediately and require a considerable amount of time to construct the solution. In order to enhance the generalization capability of language models, and as a parallel to human behavior, we propose to use special 'thinking tokens' which allow the model to perform much more calculations whenever a complex problem is encountered.
\end{abstract}

% \begin{figure}[htbp]
% \centering
% \includegraphics[width=\linewidth]{ThinkingTokens.png}
% \caption{Illustration of 'thinking tokens' (marked as $<$T$>$) in a sentence which requires a complex calculation and the positive impact of this approach on perplexity of the model (lower is better).}
% \label{fig:example}
% \end{figure}

\section{Introduction}
\vspace{-2pt}
Language models based on neural networks have gained a great deal of interest in recent years \cite{roose2022, naughton2023}. Their impressive and coherent answers amazed people across many industries. However, it has soon been discovered that these language models have problems with complex tasks \cite{gowers2023, frieder2023}. \par
Complex questions such as \textit{'how much is 56 times 37'} which are computationally requiring, are problematic for the language model to process or even answer correctly. One can argue that humans also cannot perform this calculation right away and require a considerable amount of time to provide a solution. \par

\begin{wrapfigure}[12]{r}{0.5\linewidth}
    \vspace{-30pt}
  \begin{center}
   \includegraphics[width=\linewidth]{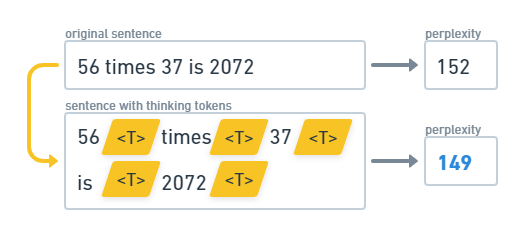}
  \end{center}
  \vspace{-15pt}
  \caption{Illustration of 'thinking tokens' (marked as $<$T$>$) in a sentence which requires a complex calculation and the positive impact of this approach on perplexity of the model (lower is better). \label{fig:example}}
\end{wrapfigure}

Although these reasoning abilities could be improved by employing a large amount of supervision by providing labeled examples to the model \cite{frieder2023}, we aim for a much faster unsupervised approach.\par 
To enhance the generalization capability of language models, we propose to use special 'thinking tokens' which allow the model to perform much more calculations whenever a complex problem is encountered. This could result in an improved generalization capability of language models, which could adapt to more complex tasks and even decide themselves what strategy is most beneficial for an encountered problem.\par

\vspace{-5pt}
\section{Related work}
\vspace{-2pt}
Research regarding reasoning can be traced back to 1959 \cite{samuel1959}, and now continues to be a big part of theorem proving \cite{urban2008,urban2017, irving2016}. Presently, large language models are being used to learn reasoning from natural language \cite{chowdhery2022, lewkowycz2022}. \par
A similar problem has also been studied in \cite{mikolov_variable}, where a language model recomputes only a part of the recurrent hidden layer. Another work with a similar motivation explores the possibility of using a neural network that is capable of learning algorithms \cite{mikolov_infering}. \par

\vspace{-5pt}
\section{Thinking tokens for language models}
\vspace{-2pt}
Our approach is to introduce special 'thinking tokens' ($<T>$) after each word in a sentence whenever a complex problem is encountered. The core idea is that each 'thinking token' would buy more time for the model before an answer is expected, which would be used to run additional computations to better answer a complex problem that was presented. This concept has a great potential in recurrent neural networks \cite{Elman1990, hochreiter1997} due to their architecture, because it enables the RNN to perform multiple in-memory operations in a single step, meaning that extra calculations can be run in the hidden layer multiple times.\par
As a proof of concept, we have added $N$ 'thinking tokens' ($<T>$) after each observed word in a dataset. Our vision is that this basic concept can be extended to a self-adjusting model, which will be able to decide itself if and how many 'thinking tokens' will be used for a specific problem, where $N$ could also vary throughout the sentence. This would allow us to reduce the computational time, which would not increase $N$ times.
The visualization of our core idea, which we aim to validate in this paper, is presented in Figure \ref{fig:example}.\par 

\vspace{-5pt}
\section{Results}
\vspace{-2pt}
Experiments execution has successfully produced numerous examples where the usage of 'thinking tokens' leads to an improvement in the model's judgment. 
Preliminary results show that sentences that require non-trivial reasoning, have the biggest improvement in perplexity when 'thinking tokens' are used compared to the standard model. This is also observable on the sample sentences from Maths dataset in Table \ref{table:first}. A larger scope of examples across all the datasets is in Appendix \ref{experiments}.\par
We can observe that introduction of 'thinking tokens' is also successful for sentences that include specific numbers or representative symbols of numerical values.\par

\begin{table}[ht]
  \renewcommand{\thetable}{\arabic{table}}
  \centering
    \small
    \begin{tabular*}{0.95\linewidth}{p{2.0cm}|p{6.0cm}|p{2.0cm}|p{2.0cm}}
    \toprule
     \bfseries Dataset & \bfseries Sentence & \bfseries Ppl. orig. $\downarrow$ & \bfseries Ppl. $<$T$>$ $\downarrow$ \\
     \midrule
        maths & What is the remainder when 8922293 is divided by 263 ? 18 & 16.8 & 13.1 \\
        \midrule
        maths & Convert -464 (base 9) to base 6 . -1434 & 24.3 & 19.8 \\
        \bottomrule
  \end{tabular*}
  \caption{Examples of sentences where the introduction of 'thinking tokens' is beneficial to the model. First and second column refer to the name of the dataset and the specific sentence. Two rightmost columns refer to the perplexity without tokens (Ppl. orig.) and with 'thinking tokens' (Ppl. $<$T$>$). A downward arrow $\downarrow$ indicates that lower is better.}\label{table:first}
\end{table}
\vspace{-17pt}

\section{Discussion and Future work}
\label{future}
\vspace{-2pt}
Language models often make mistakes in complex problems like calculations or reasoning, since they rely on large training sets and their great memorization capability. We show that giving RNNLM more time to 'think' and not pressuring the model to produce an answer immediately, helps the model resolve various complex tasks more accurately.\par
Building on the proof of concept, we plan to extend our research and create a model that would be able to decide itself how much extra time is needed in order to produce the best answer possible. If successful, this concept could be implemented as a default behavior for language models that encounter complex and computationally demanding tasks. We also believe that the ability of a model to self-regulate this factor would vastly improve adaptability and generalization capability of language models in general.\par

% \section{Acknowledgements}

% This work was supported by NSF grant No.\ PHY-9723972.

\footnotesize

\bibliographystyle{plain}
\bibliography{aitp2022}

\newpage

\appendix
\section{Appendix}
\label{Appendix}

\subsection{Experiments}
\label{experiments}
\begin{table}[h]
  \renewcommand{\thetable}{\arabic{table}}
  \centering
    \small
    \begin{tabular*}{0.8\linewidth}{p{1.5cm}|p{6.0cm}|p{1.2cm}|p{1.5cm}}
    \toprule
     \bfseries Dataset & \bfseries Sentence & \bfseries Ppl. orig. $\downarrow$ & \bfseries Ppl. $<$T$>$ $\downarrow$ \\
     \midrule
        ptb & britain has two main index-arbitrage instruments & 24.1 & 21.7 \\
        \midrule
        ptb & today pc shipments annually total some N billion world-wide & 45.5 & 42.3 \\
        \midrule
        wt-2 & The discography of LiSA includes three studio albums , one extended play , ten singles , and five video albums & 130.1 & 127.7 \\
        \midrule
        wt-2 & Complex N lies to the west of the Bat Palace and Temple III . The complex dates to AD 711 & 246.8 & 244.2 \\
        \midrule
        etb & increase in deficit raises the interest rate & 20.5 & 18.5 \\
        \midrule
        etb & however too much money in circulation can lead to inflation & 27.9 & 25.5 \\
        \bottomrule
  \end{tabular*}
  \caption{Examples of sentences where the introduction of 'thinking tokens' is beneficial to the model. First and second column refer to the name of the dataset and the specific sentence. Two rightmost columns refer to the perplexity without tokens (Ppl. orig.) and with 'thinking tokens' (Ppl. $<$T$>$). A downward arrow $\downarrow$ indicates that lower is better.}\label{table:second}
\end{table}

To evaluate the plausibility of our idea, we first propose to extend the standard recurrent neural language model with extra tokens. This does not require any change in the model architecture and can be achieved by modifying the input data by adding $N$ 'thinking tokens' after each word. In our case $N$ $=$ 1. \par
We have chosen a simple setup where a RNN LM with one hidden layer is used. We train a baseline model, standard LSTM LM \cite{hochreiter1997}, and finally a model with the 'thinking tokens', as we believe the perplexity differences could be rather small. After all, the mistakes in reasoning about numbers influence entropy much less than for example correctly capturing uni-gram frequencies of the most common words. \par
We have designed an experiment to identify sentences in which the largest differences in perplexity between the two models can be observed. This could allow us to determine in which cases the usage of 'thinking tokens' is beneficial for the model. For the purpose of fair results evaluation, the loss generated by 'thinking tokens' is omitted from the calculation of perplexity. \par
Models were trained on standard language modeling tasks like Penn TreeBank \cite{mikolov2010}, WikiText-2 \cite{smerity2017} and also on mathematics dataset \cite{Saxton2019} and dataset retrieved from MacroEconomics textbook \cite{cooper_john_2012}. Hyper-parameters and additional experiments are listed in \ref{model_hyper_params}. \par

\subsection{Word probabilities}
To give the reader more insight into what happens when the 'thinking token' is used, we have decided to show the probabilities for each word in two sample sentences. It is important to note that the probabilities of 'thinking tokens' are omitted. \par

\textit{'What is the remainder when 8922293 is divided by 263 ? 18'}. \\
\\
Word: What \\
LSTM: 0.27570505869594907016 \\
LSTM+$<$T$>$: 0.27460685731534064 \\
Word: is \\ 
LSTM: 0.9983407855033875 \\
LSTM+$<$T$>$: 0.9994571208953857 \\
Word: the \\
LSTM: 0.7941651940345764 \\
LSTM+$<$T$>$: 0.7810274958610535 \\
Word: remainder \\
LSTM: 0.05905058979988098 \\
LSTM+$<$T$>$: 0.39061781764030457 \\
Word: when \\
LSTM: 0.9977582693099976 \\
LSTM+$<$T$>$: 0.9991976022720337 \\
Word: 8922293 \\
LSTM: 8.769490705162752e-06 \\
LSTM+$<$T$>$: 2.8874606869067065e-05 \\
Word: is \\
LSTM: 0.9854738712310791 \\
LSTM+$<$T$>$: 0.977165162563324 \\
Word: divided \\
LSTM: 0.9997884631156921 \\
LSTM+$<$T$>$: 0.9993095993995667 \\
Word: by \\
LSTM: 0.9998854994773865 \\
LSTM+$<$T$>$: 0.9999291300773621 \\
Word: 263 \\
LSTM: 3.43535648426041e-05 \\
LSTM+$<$T$>$: 4.3290932808304206e-05 \\
Word: ? \\
LSTM: 0.9997261762619019 \\
LSTM+$<$T$>$: 0.9896721243858337 \\
Word: 18 \\
LSTM: 0.02015523798763752 \\
LSTM+$<$T$>$: 0.019233182072639465 \\

\textit{'Increase in deficit raises the interest rate'}. \\
\\
Word: increase \\
LSTM: 0.00013779969594907016 \\
LSTM+$<$T$>$: 0.0005539595731534064 \\
Word: in
LSTM: 0.7479172348976135 \\
LSTM+$<$T$>$: 0.723701536655426 \\
Word: deficit
LSTM: 0.0001904059899970889 \\
LSTM+$<$T$>$: 0.00011684149649227038 \\
Word: raises
LSTM: 0.01803828589618206 \\
LSTM+$<$T$>$: 0.004907318856567144 \\
Word: the
LSTM: 0.46662768721580505 \\
LSTM+$<$T$>$: 0.4531695246696472 \\
Word: interest
LSTM: 0.05833666771650314 \\
LSTM+$<$T$>$: 0.07482223212718964 \\
Word: rate
LSTM: 0.9725480079650879 \\
LSTM+$<$T$>$: 0.9811777472496033 \\

\subsection{Perplexity of models}
\label{ppl_models}

\begin{table}[htbp]
    \centering
    \begin{tabular}{c|c|c}
    \multicolumn{3}{c}{\textbf{Validation perplexity}} \\
    \hline
     \bfseries Dataset & \bfseries LSTM& \bfseries LSTM+$<$T$>$ \\
     \hline
       \textit{Penn tree bank}  & 68.2 & 68.4 \\
       \hline
       \textit{Wikitext-2}   & 76.8 & 82.4 \\
       \hline
       \textit{Economic textbooks}   & 49.0 & 51.4 \\
       \hline
       \textit{Maths}   & 19.8 & 19.8 \\
       
    \end{tabular}
    \caption{4 datasets ptb \cite{mikolov2010}, wt-2 \cite{smerity2017}, etb \cite{cooper_john_2012}, maths \cite{Saxton2019}.}

\label{table:ppl_models}
\end{table}
In Table \ref{table:ppl_models} we have evaluated language models on 4 datasets. In the first column, we have standard LSTM with 1 layer while in the second column, we have results for the same LSTM, but with a 'thinking token' after each observed word. It is important to note that loss from 'thinking tokens' was not included in the calculation of perplexity. \par
It could be observed that addition of 'thinking token' results in slight performance decrease in perplexity. However, the main goal of $<T>$ tokens is not to improve perplexity, but to enhance model's capability to 'think'.\par

\subsection{Number of thinking tokens}
\label{n_o_thinking_tokens}

\begin{table}[htbp]
    \centering
    \begin{tabular}{c|c|c}
    \multicolumn{3}{c}{\textbf{Validation perplexity}} \\
    \hline
     \bfseries Dataset & \bfseries LSTM+2$<$T$>$& \bfseries LSTM+3$<$T$>$ \\
     \hline
       \textit{ptb}  & 71.3 & 78.6 \\
       \hline
       \textit{wt-2}   & 89.1 & 94.0 \\
       \hline
       \textit{etb}   & 56.2 & 60.9 \\
       \hline
       \textit{maths}   & 21.0 & 24.4 \\
       
    \end{tabular}
    \caption{Evaluation of validation perplexity when number of 'thinkings tokens' differs}

\label{table:n_o_tokens}
\end{table}
We have also investigated how the number of 'thinking tokens' influences the result as shown in Table \ref{table:n_o_tokens}. \par 
Adding more 'thinking tokens' worsens the validation perplexity with LSTM model. Explanation of this trend could be that using more 'thinking tokens' is not always beneficial, since it increases the chance of a model to forget what was before 'thinking tokens'. 

\subsection{Model hyper-parameters}
\label{model_hyper_params}
\begin{table}[htbp]
    \centering
    \begin{tabular}{c|c}
    \multicolumn{2}{c}{\textbf{Hyper-parameters}} \\
    \hline
     \bfseries Parameter & \bfseries Value\\
     \hline
       \textit{Bptt}  & 70 \\
       \hline
       \textit{Batch size}   & 12  \\
       \hline
       \textit{Gradient clipping}   & 0.25 \\
       \hline
       \textit{Hidden neurons}   & 450 \\
       \hline
       \textit{Layers}   & 1 \\
       \hline
       \textit{Hidden neurons}   & 450 \\

    \end{tabular}
    \caption{Model hyper-parameters}

\label{table:hyper_params}
\end{table}
Hyper-parameters used to train our LSTM are listed in Table \ref{table:hyper_params}.
We have used the ASGD trick in training \cite{smerity2017}.

\end{document}